\documentclass{article}

\usepackage[final]{neurips_2025}

\usepackage[utf8]{inputenc} % allow utf-8 input
\usepackage[T1]{fontenc}    % use 8-bit T1 fonts
\usepackage{hyperref}       % hyperlinks
\usepackage{url}            % simple URL typesetting
\usepackage{booktabs}       % professional-quality tables
\usepackage{amsfonts}       % blackboard math symbols
\usepackage{nicefrac}       % compact symbols for 1/2, etc.
\usepackage{microtype}      % microtypography
\usepackage{xcolor}         % colors
\usepackage{amsthm}      % quinn
\usepackage{amsmath}     % quinn
\usepackage{multirow}    % quinn
\usepackage{paralist}    % quinn
\usepackage{graphicx}    % quinn
\usepackage{cleveref}    % quinn
\usepackage{algorithm}   % quinn
\usepackage{algpseudocode} % quinn
\usepackage{tablefootnote} % quinn
\usepackage{subcaption}  % quinn

\newtheorem*{remark*}{Remark}
\newtheorem{proposition}{Proposition}

\title{Continuous Subspace Optimization \\ for Continual Learning}

\author{
    Quan Cheng\textsuperscript{\rm 1,2},~~Yuanyu Wan\textsuperscript{\rm 3,4},~~Lingyu Wu\textsuperscript{\rm 1,2},~~Chenping Hou\textsuperscript{\rm 5},~~Lijun Zhang\textsuperscript{\rm 1,2,}\thanks{Lijun Zhang is the corresponding author.}\\
    \textsuperscript{\rm 1}National Key Laboratory for Novel Software Technology, Nanjing University, Nanjing, China \\
    \textsuperscript{\rm 2}School of Artificial Intelligence, Nanjing University, Nanjing, China \\
    \textsuperscript{\rm 3}School of Software Technology, Zhejiang University, Ningbo, China \\
    \textsuperscript{\rm 4}Hangzhou High-Tech Zone (Binjiang) Institute of Blockchain and Data Security, Hangzhou, China \\
    \textsuperscript{\rm 5}College of Science, National University of Defense Technology, Changsha, China \\
    \texttt{\{chengq, wuly, zhanglj\}@lamda.nju.edu.cn}\\
    \texttt{wanyy@zju.edu.cn},~~\texttt{hcpnudt@hotmail.com}
}

\begin{document}

\maketitle

\begin{abstract} 
    Continual learning aims to learn multiple tasks sequentially while preserving prior knowledge, but faces the challenge of catastrophic forgetting when adapting to new tasks. Recently, approaches leveraging pre-trained models have gained increasing popularity in mitigating this issue, due to the strong generalization ability of foundation models. To adjust pre-trained models for new tasks, existing methods usually employ low-rank adaptation, which restricts parameter updates to a fixed low-rank subspace. However, constraining the optimization space inherently compromises the model's learning capacity, resulting in inferior performance. To address this limitation, we propose \textbf{\underline{Co}}ntinuous \textbf{\underline{S}}ubspace \textbf{\underline{O}}ptimization for Continual Learning (CoSO) to fine-tune the model in a series of subspaces rather than a single one. These sequential subspaces are dynamically determined through the singular value decomposition of the gradients. CoSO updates the model by projecting gradients onto these subspaces, ensuring memory-efficient optimization. To mitigate forgetting, the optimization subspace of each task is constrained to be orthogonal to the historical task subspace. During task learning, CoSO maintains a task-specific component that captures the critical update directions for the current task. Upon completing a task, this component is used to update the historical task subspace, laying the groundwork for subsequent learning. Extensive experiments on multiple datasets demonstrate that CoSO significantly outperforms state-of-the-art methods, especially in challenging scenarios with long task sequences.
\end{abstract}

\section{Introduction}

    Deep neural networks have achieved remarkable success when trained on large-scale offline data under the assumption of independent and identically distributed (i.i.d.) samples \citep{He2016resnet,Vaswani2017transformer,dosovitskiy2021vit}. However, real-world applications often require models to learn from a sequence of tasks with different data distributions, a scenario known as continual learning \citep{delange2021clsurvey,van2022three,Masana2022class,wang2024continualsurvey,zhou2024class}. The major challenge in continual learning is catastrophic forgetting \citep{MCCLOSKEY1989catastrophic}, where the model's performance on previously learned tasks deteriorates significantly as it adapts to new tasks. 

    In recent years, pre-trained models especially vision transformers (ViTs) \citep{dosovitskiy2021vit} have demonstrated exceptional performance across various downstream tasks through their robust generalization ability. This property makes pre-trained models highly promising in mitigating catastrophic forgetting, leading to a growing research focus on continual learning with foundation models \citep{Smith2023coda,lu2024visual, Liang2024inflora, Zhou2024continual,wu2025sdlora}. To efficiently fine-tune pre-trained ViTs, existing continual learning methods \citep{gao2023lae,Liang2024inflora,wu2025sdlora} employ low-rank adaptation (LoRA) \citep{hu2022lora} to optimize the models, which confine parameter updates to a specific low-rank subspace to reduce the interference between tasks. However, this rigid constraint on update directions inherently limits the model's learning capacity, leading to inferior performance.
    
    To address this issue, we propose \textbf{Co}ntinuous \textbf{S}ubspace \textbf{O}ptimization for Continual Learning (CoSO), which achieves enhanced adaptability by optimizing the model within multiple subspaces rather than a fixed one. These sequential subspaces are derived from the singular value decomposition of the gradients. By projecting gradients onto these low-dimensional subspaces for Adam \citep{kingma2014adam} optimization and then projecting back for parameter updates, CoSO achieves memory-efficient learning. To prevent forgetting, we enforce orthogonality between the optimization subspaces of current and historical tasks during training. While learning a task, CoSO leverages Frequent Directions (FD) \citep{Ghashami2016fd,wan2018fd,wan2022fd} to maintain a compact task-specific component, which captures critical update directions of the current task with negligible computational cost. After completing the current task, we use this dedicated component to estimate the task-specific subspace, which is then integrated into the historical task subspace, laying the groundwork for subsequent learning.
    
    Experimental results on CIFAR100, ImageNet-R, and DomainNet show that CoSO consistently outperforms state-of-the-art methods by a significant margin across diverse continual learning settings, particularly in challenging scenarios involving long task sequences. The substantial performance gains highlight CoSO's strong potential for real-world continual learning.

    In summary, our contributions are as follows:
    \begin{compactitem}
    \item We propose CoSO, a novel continual learning framework that fine-tunes pre-trained models via continuous gradient-derived subspaces, enabling efficient adaptation to sequential tasks.
    \item We introduce a lightweight mechanism to maintain the historical task subspace, enabling CoSO to keep current updates orthogonal to the historical subspace and thereby mitigate task interference.
    \item We conduct extensive experiments, demonstrating CoSO's superior performance over prior PEFT-based continual learning methods across various datasets and settings.
    \end{compactitem}

\section{Related Work}

    In this section, we review related work on continual learning and low-rank optimization in offline learning. 

\subsection{Continual Learning}

    Continual learning \citep{delange2021clsurvey,van2022three,Masana2022class,wang2024continualsurvey,zhou2024class} aims to enable neural networks to incrementally learn from a sequence of tasks while retaining previously learned knowledge. These approaches broadly fall into five categories \citep{wang2024continualsurvey}: regularization-based methods \citep{zenke2017si,kirkpatrick2017ewc,li2017lwf}, replay-based methods \citep{Lopez-Paz2017gem,rebuffi2017icarl,chaudhry2018agem,chaudhry2019tiny,liu2020mnemonics,sun2022exploring}, optimization-based methods \citep{farajtabar2020orthogonal,saha2021gpm,wang2021nullspacecl}, representation-based methods \citep{madaan2022representational,pham2024dualnet}, and architecture-based methods \citep{yoon2018den,li2019learngrow,sokar2021spacenet,liang2023duaigpm}. Regularization-based methods introduce additional loss terms to constrain parameter updates, preventing drastic changes in parameters that are important for early tasks. Replay-based methods store a small subset of training samples from previous tasks in a limited buffer and periodically replay these samples alongside new data, allowing the model to rehearse earlier knowledge. Optimization-based methods manipulate the update directions of each task according to preserved information of previous tasks. Representation-based methods utilize statistical information of features to calibrate classifiers. Architecture-based methods dynamically modify network architectures, dedicating specific model capacity for new tasks.

    Early continual learning approaches typically initialize their models with random weights. The strong generalization capabilities of foundation models, especially vision transformers \citep{dosovitskiy2021vit}, have made pre-trained architectures an increasingly attractive solution for continual learning \citep{Zhou2024continual}. Recent developments in parameter-efficient fine-tuning (PEFT) based continual learning methods \citep{gao2023lae,Liang2024inflora,lu2024visual,wu2025sdlora} have facilitated efficient adaptation of foundation models through selective parameter optimization, substantially lowering computational requirements. Existing PEFT-based methods can be broadly categorized into two groups: (1) prompt-based techniques that focus on optimizing learnable tokens \citep{lester2021power,wang2022dualprompt,wang2022learning,Smith2023coda,lu2024visual}, and (2) LoRA-based methods that adjust parameters within constrained low-dimensional subspaces \citep{gao2023lae,Liang2024inflora,wu2025sdlora}. 

    Among prompt-based approaches, L2P \citep{wang2022learning} introduces task-specific prompt tokens to modulate the pre-trained model's behavior, but struggles with knowledge transfer between tasks. DualPrompt \citep{wang2022dualprompt} addresses this limitation by maintaining both task-specific and task-invariant prompts, enabling better knowledge sharing. CODA-Prompt \citep{Smith2023coda} further enhances adaptation flexibility through dynamic prompt composition from a shared pool. $\text{VPT-NSP}^2$ \citep{lu2024visual} learns each task by tuning learnable prompts in the null space of previous tasks' features. However, these methods influence model behavior indirectly through learnable tokens, which restrict the model's ability to capture complex task-specific features.

    Complementary to prompt-based methods, LoRA-based approaches directly update model parameters in a parameter-efficient manner. InfLoRA \citep{Liang2024inflora} constrains the parameter updates within a predetermined subspace to reduce the interference between tasks. SD-LoRA \citep{wu2025sdlora} decouples the learning of the magnitude and direction of LoRA components. However, both methods confine weight updates to a specific low-rank subspace, which inherently limits the model's learning capacity. Unlike these methods, CoSO updates the parameters across a series of subspaces, enabling the learning of full-rank weights and thereby enhancing the model's flexibility.

\subsection{Low-rank Optimization in Offline Learning}

    Low-rank adaptation (LoRA) \citep{hu2022lora} has gained significant attention for its ability to reduce computational and memory requirements when fine-tuning pre-trained models \citep{mao2022unipelt,zhang2023adalora}. Specifically, LoRA reparameterizes the update of a linear layer's weights $\Delta W = BA \in \mathbb{R}^{m\times n}$, where $B\in \mathbb{R}^{m\times r}, A\in \mathbb{R}^{r\times n}$ are low-rank matrices. By freezing the original weights and only updating the low-rank components, LoRA enables parameter-efficient fine-tuning while preserving performance in many downstream tasks. However, it has been demonstrated \citep{xia2024chain} that low-rank weight updates limit the performance compared to full-rank fine-tuning. Recent works \citep{Cosson2023lowrankgrad, Zhao2024galore} have shown that neural network gradients often exhibit low-rank structure. Instead of approximating the weight matrix as low rank, GaLore \citep{Zhao2024galore} directly leverages the low-rank gradients to optimize the model. This methodology enables memory-efficient optimization through effective dimensionality reduction in gradient spaces.

    To be concrete, GaLore utilizes the singular value decomposition (SVD) of $G_t\in \mathbb{R}^{m \times n}$ to compute a low-rank projection matrix $P_t \in \mathbb{R}^{m \times r}$, where $r \ll n$ is the target rank. Leveraging $P_t$, GaLore transforms the gradient $G_t$ into a compact form $P_t^\top G_t$ to achieve memory-efficient parameter updates. At each training step $t$, the gradient update can be decomposed into three operations:
    \begin{align}
        R_t &= P_t^\top G_t & \text{(forward projection)} \notag \\
        N_t &= \text{Adam}(R_t) & \text{(adam optimizer update)} \notag \\
        \tilde{G}_t&= P_t N_t. & \text{(backward projection)} \notag
    \end{align}
    The projection matrix $P_t$ is periodically updated through SVD to follow the evolving gradient subspace. Utilizing the final gradient $\tilde{G}_t$, GaLore updates the model parameters with learning rate $\eta$:
    \begin{equation}
        W_t = W_{t-1} - \eta \tilde{G}_t. \notag
    \end{equation}
    Compared to LoRA, GaLore not only reduces memory storage from $(mn + 3mr + 3nr)$ to $(mn + mr + 2nr)$, but also achieves higher model capacity by directly optimizing in the most relevant gradient subspaces rather than constraining updates to a predefined low-rank structure.

\section{Methodology}
\label{sec:method}

    In this section, we first introduce the necessary preliminaries, then present the details of our approach.

\subsection{Preliminaries}

    In continual learning, a model needs to learn a sequence of tasks while retaining knowledge of previous tasks. We consider the class-incremental learning setting, where task identities are unavailable at inference time and access to historical data is prohibited during learning new tasks \citep{wang2022dualprompt,Smith2023coda,Liang2024inflora,lu2024visual,wu2025sdlora}. We denote the task sequence as $\mathcal{D} = \{\mathcal{D}_1, ..., \mathcal{D}_N\}$, where each task dataset $\mathcal{D}_\tau = \{(\mathbf{x}_{i,\tau}, y_{i,\tau})\}_{i=1}^{n_\tau}$ contains $n_\tau$ input-label pairs. Following recent work \citep{wang2022dualprompt,gao2023lae}, we adopt a pre-trained Vision Transformer (ViT) \citep{dosovitskiy2021vit} as the backbone network, denoted as $f_\Theta(\cdot)$ with parameters $\Theta$, and classifier $h_\Phi(\cdot)$ with parameters $\Phi$, thus the model is $h_\Phi(f_\Theta(\cdot))$. Formally, the hidden state $Y^\ell_{\tau}$ of feature $X^\ell_{\tau}$ at the linear layer $\ell$, can be calculated as $Y^\ell_{\tau} = W^\ell X^\ell_{\tau}$, where $W^\ell$ is the weight matrix of the linear layer. Let $G_{\tau, t}^{\ell}$ denote the gradient at the $t$-th training step of the linear layer $\ell$ in task $\tau$. For simplicity, we omit the symbol $\ell$, using $W$ to refer $W^\ell$ and $G_{\tau, t}$ to refer $G_{\tau, t}^{\ell}$ in the following sections.

\subsection{Continuous Subspaces Optimization}
\label{sec:32}

\begin{figure}
        \centering
        \includegraphics[width=\textwidth]{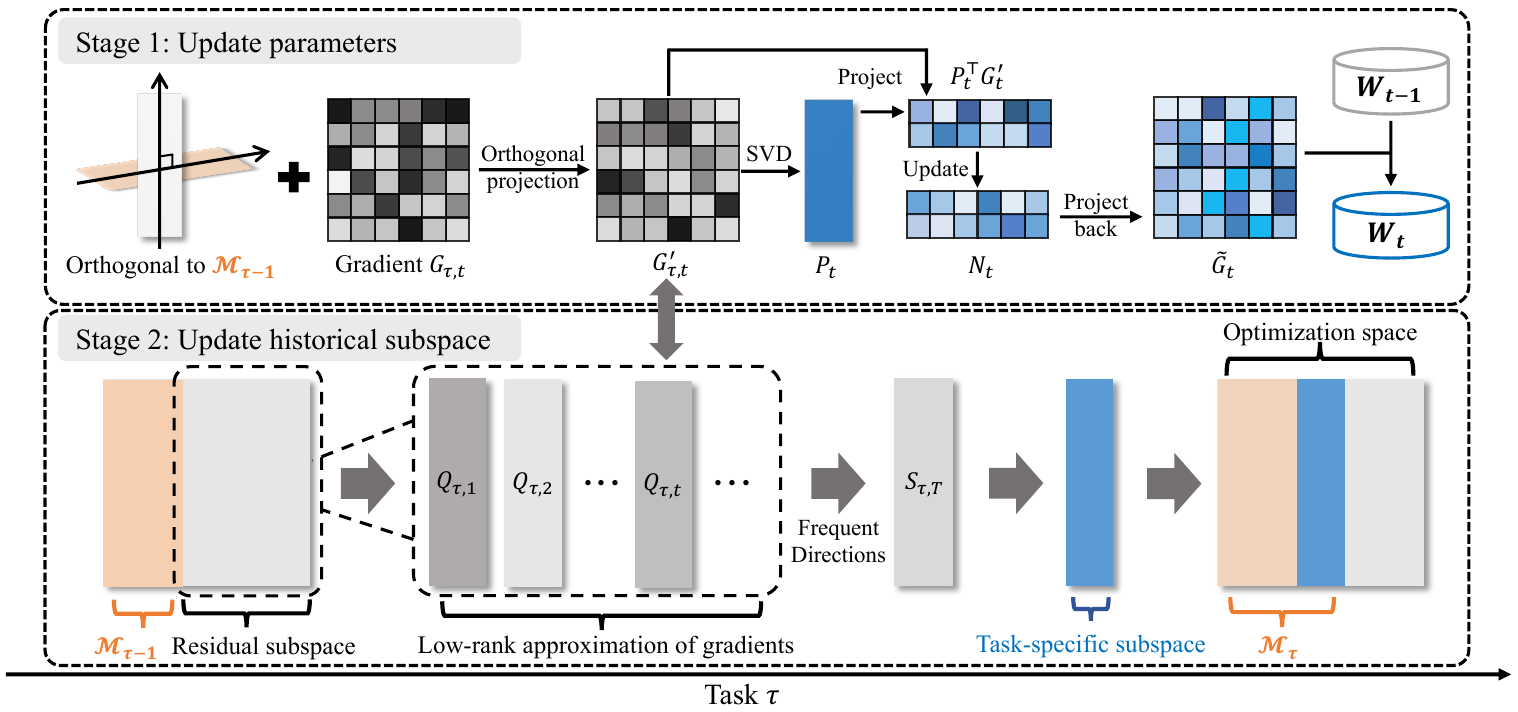}
        \caption{CoSO optimizes the parameters in continual low-rank subspaces, enhancing the learning capacity of models. To mitigate forgetting, the optimization subspaces of the current task are set to be orthogonal to the historical task subspace. While learning a task, CoSO consolidates the low-rank approximation matrices $\{Q_{\tau,t}\}_{t=1}^{T}$ into a task-specific component $S_{\tau,T}$ through Frequent Directions. The dedicated component is then used to update the historical task subspace spanned by $\mathcal{M}_{\tau-1}$.}
        \label{fig:coso}
    \end{figure}

    Inspired by GaLore~\citep{Zhao2024galore}, we propose CoSO to address the rigidity of single subspace adaptation methods through multiple subspaces optimizing. However, directly using GaLore causes severe interference between different tasks in continual learning. To minimize the interference, CoSO enforces orthogonal constraints between current and historical subspace during training. Motivated by memory consolidation in cognitive neuroscience~\citep{Dudai2004neurobiology}, CoSO estimates a task-specific subspace to consolidate knowledge upon learning each task. This subspace preserves critical learning directions of the task based on gradients at all training steps, and is incrementally integrated into the historical task subspace, enabling efficient knowledge accumulation. The whole process of CoSO is illustrated in Figure~\ref{fig:coso}. We first introduce how to optimize the model in continuous subspaces in this section. Then we present how to update the historical task subspace in Section~\ref{method:subspace}.

    In continual learning, the key challenge is to prevent new task updates from interfering with previously learned knowledge. Building on the insight that gradient updates in neural networks typically lie in the span of input features~\citep{saha2021gpm}, we develop an approach that leverages this gradient-input feature relationship to minimize task interference through orthogonal projection. Specifically, we maintain an orthogonal basis matrix $\mathcal{M}_{\tau-1}$ that spans the gradient subspace accumulated from all previous tasks prior to the current task $\tau$. Since gradients inherently encode information about the input features they were computed from, this historical subspace captures the principal directions that were important for learning previous tasks. We recognize that gradient steps along these historical directions would cause maximal interference with past learning, while gradient steps orthogonal to this space result in minimal interference. For each gradient $G_{\tau, t}$ computed during training on task $\tau$, we project it onto the orthogonal complement of historical subspace:
    \begin{equation}
        G_{\tau, t}^{\prime} = G_{\tau, t} - \mathcal{M}_{\tau-1}\mathcal{M}_{\tau-1}^\top G_{\tau, t}.
    \end{equation}

    This projection removes the gradient component aligned with the learning directions of previous tasks, leaving only the orthogonal component $G_{\tau, t}^{\prime}$ for updating the parameters. When we update the weight matrix using these orthogonal gradients, i.e., $\Delta W = - \eta \sum_{t=1}^{T} G_{\tau, t}^{\prime}$, the parameter changes occur in directions that have minimal overlap with the optimization trajectories of previous tasks. This approach effectively partitions the parameter space, preserving directions important for past tasks while utilizing orthogonal directions for new learning. The orthogonal projection thus provides a principled way to balance plasticity for new tasks with stability for old tasks, enabling the model to expand its capabilities while mitigating interference.
    %%%%%%

    However, updating the model with the full orthogonal gradient $G_{\tau,t}^{\prime}$ incurs substantial memory overhead and high computational cost, particularly in vision transformers. To achieve memory-efficient fine-tuning, we follow GaLore \citep{Zhao2024galore} and decompose $G_{\tau, t}^\prime \in \mathbb{R}^{m\times n}$ using singular value decomposition (SVD) to get the projection matrix $P_{\tau, t}$:
    \begin{equation}
    \label{eq:project-matrix}
    \begin{aligned}
        & U\Sigma V^\top = \text{SVD}_{r_1}(G_{\tau, t}^\prime) \\
        & P_{\tau, t} = U[:,:r_1],
    \end{aligned}
    \end{equation}
    where $m$ and $n$ are the dimensions of the original weight matrix, $r_1 \ll n$ is the target projection rank, and $\text{SVD}_{r_1}(\cdot)$ denotes a truncated SVD that retains the top-$r_1$ singular values. Subsequently, we project the orthogonalized gradient $G_{\tau, t}^\prime$ into the low-rank subspace spanned by $P_{\tau, t}$, effectively reducing the memory footprint of parameter updates:
    \begin{equation}
        R_{\tau, t} = P_{\tau, t}^\top G_{\tau, t}^\prime.
    \end{equation}
Then $R_{\tau, t}$ is updated by Adam \citep{kingma2014adam} as follows:
\begin{equation}
\label{eq:adam-update}
    \begin{aligned}
        M_{\tau, t} &= \left(\beta_1 \cdot M_{\tau,t-1} + (1 - \beta_1) \cdot R_{\tau, t} \right) / (1 - \beta_1^t)\\
        V_{\tau, t} &= \left( \beta_2 \cdot V_{\tau,t-1} + (1 - \beta_2) \cdot R_{\tau, t}^2 \right) / (1 - \beta_2^t)\\
        N_{\tau, t} &= M_{\tau, t} / \left(\sqrt{V_{\tau, t}} + \epsilon \right),
    \end{aligned}    
\end{equation}
where $\beta_1, \beta_2$ are decay rates, $M_{\tau, t}$ is the first-order momentum, and $V_{\tau, t}$ is the second-order momentum. The low-rank normalized gradient $N_{\tau, t}$ is then projected back to update the parameters with learning rate $\eta$:
\begin{equation}
\label{eq:update}
    \begin{aligned}
        \tilde{G}_{\tau, t} &= P_{\tau, t} N_{\tau, t} \\
        W_{\tau, t} &= W_{\tau, t-1} - \eta \cdot \tilde{G}_{\tau, t}.
    \end{aligned}
\end{equation}

Because $P_{\tau,t}$ is computed from the projected gradient $G_{\tau, t}^{\prime}$, which is orthogonal to the historical subspace spanned by $\mathcal{M}_{\tau-1}$, any parameter updates derived from $P_{\tau,t}$ remain in the null space of previous tasks' feature spaces. Consequently, the linear layer's output for every earlier task remains unchanged, preventing interference at the representation level. Since $P_{\tau, t}$ is dynamically changed to capture the most important directions of $G_{\tau, t}^\prime$, we are optimizing the model in continuous subspaces rather than a fixed one, thereby expanding the model's representational adaptability. To balance computational efficiency, we update the projection matrix $P_{\tau,t}$ every $K$ steps. By updating $R_{\tau, t}$ in lower dimension space, the memory requirement is reduced from $(mn + 3mr_1 + 3nr_1)$ to $(mn + mr_1 + 2nr_1)$ compared to LoRA-based methods, such as InfLoRA \citep{Liang2024inflora} and SD-LoRA \citep{wu2025sdlora}.

\subsection{Historical Task Space Update}
\label{method:subspace}

\textbf{Task-Specific Subspace Estimation.} To update the orthogonal basis matrix $\mathcal{M}_{\tau-1}$ of the historical task space, we need to efficiently estimate a task-specific subspace, which retains the critical gradient directions of the current task. Specifically, for task $\tau$, the model undergoes $T$ training steps, producing a sequence of gradients $\{G_{\tau,1}^\prime,...,G_{\tau,T}^\prime\}$, where $G_{\tau,t}^\prime \in \mathbb{R}^{m\times n}$. To identify the primary directions of these gradients, we consider the accumulated covariance matrix $\sum_{t=1}^T G_{\tau,t}^\prime G_{\tau,t}^{\prime\top} \in \mathbb{R}^{m \times m}$, which integrates information from all training steps and characterizes the subspace where most updates occur. However, directly maintaining such accumulated covariance matrix would be computationally expensive, requiring $O(m^2nT)$ time complexity. This is particularly challenging for transformer-based models where the parameter dimension $m, n$ are typically in the order of thousands. 

To ensure computational efficiency, we use Frequent Directions (FD) \citep{Ghashami2016fd,wan2018fd,wan2022fd}, a deterministic matrix sketching algorithm, to maintain a low-rank approximation of streaming gradients. The FD algorithm processes the gradients sequentially while providing a guarantee on approximation quality \citep{wan2021approxfd,yang2025dimensionfree}. Specifically, we first compute a low-rank matrix $Q_{\tau, t} \in \mathbb{R}^{m\times r_2}$ with $r_2 \ll n$ through singular value decomposition (SVD):

\begin{equation}
\label{eq:Q}
    \begin{aligned}
        & U\Sigma V^\top = \text{SVD}_{r_2}(G_{\tau,t}^\prime) \\
        & Q_{\tau,t} = U\Sigma.
    \end{aligned}
\end{equation}
Here, $\text{SVD}_{r_2}(\cdot)$ denotes a truncated SVD that retains the top-$r_2$ singular values. The resulting low-rank matrix $Q_{\tau, t}$ enables us to efficiently approximate the gradient covariance matrix:

\begin{equation}
Q_{\tau, t}Q_{\tau, t}^\top \approx G_{\tau, t}^\prime G_{\tau, t}^{\prime\top}.
\end{equation}

Based on this approximation, we further compute a sketch matrix $S_{\tau, t}\in \mathbb{R}^{m\times r_2}$ that incrementally consolidates the gradient covariance information from all training steps up to step $t$. This consolidation is achieved by combining the previous sketch matrix $S_{\tau,t-1}$ with the current approximation $Q_{\tau,t}$. The update of $S_{\tau,t}$ is as follows:

\begin{equation}
\label{eq:S}
\begin{aligned}
    & U^{\prime}\Sigma^{\prime}V^{\prime\top} = \text{SVD}_{r_2}([S_{\tau, t-1}, Q_{\tau,t}]) \\
    & S_{\tau,t} = U^{\prime}\sqrt{{\Sigma^{\prime}}^2 - \sigma_t I_{r_2}}, \sigma = {\Sigma^{\prime}}^2_{r_2,r_2}.
\end{aligned}
\end{equation}

We initialize $S_{\tau,1} = Q_{\tau,1}$, and after $T$ iterations, we obtain the final task-specific sketch matrix $S_{\tau, T}$, which satisfies:

\begin{equation}
    S_{\tau, T}S_{\tau, T}^\top \approx \sum_{t=1}^T G_{\tau,t}^\prime G_{\tau,t}^{\prime\top}.
\end{equation}

By analyzing the dominant singular vectors of $S_{\tau, T}$, we can effectively estimate the principal subspace of the current task. Note that we update $S_{\tau,t}$ every $K$ steps to match the update frequency of projection matrix $P_{\tau,t}$, ensuring consistency in our approximation process. The effectiveness of CoSO relies on the accuracy of low-rank approximation, which is formalized through the following Proposition~\ref{thm:low_rank2}.

\begin{proposition}
\label{thm:low_rank2}
Given a sequence of projected gradients $\{G'_{\tau,t}\}_{t=1}^{T}$ and low-rank matrix $\{Q_{\tau,t}\}_{t=1}^{T}$, where $G_{\tau,t}^\prime\in \mathbb{R}^{m\times n}$ and $Q_{\tau,t}\in \mathbb{R}^{m\times r_2}$. The final sketch matrix is $S_{\tau, T}\in\mathbb{R}^{m\times r_2}$. Let $A=\sum_{t=1}^T G_{\tau,t}^\prime G_{\tau,t}^{\prime\top}$, $\tilde{A} = \sum_{t=1}^T Q_{\tau,t}Q_{\tau,t}^\top$. For any $k<r_2$ the approximation error is bounded by:
\begin{equation}
\label{eq:fd_global_bound}
    \|A - S_{\tau, T}S_{\tau, T}^\top\|_2 \leq \sum_{t=1}^{T} \sigma_t^2 + \frac{\|\tilde A-[\tilde A]_k\|_{F}^{2}}{r_2 - k},
\end{equation}
where $\sigma_t$ is the $(r_2+1)$-th singular value of $G_{\tau,t}^\prime$ and $[\tilde A]_k$ is the minimizer of $\|\tilde A-[\tilde A]_k\|_{F}$ overall rank $k$ matrices.
\end{proposition}

\begin{remark*}
    Because the gradients often exhibit low-rank structure \citep{Cosson2023lowrankgrad,Zhao2024galore}, their singular values decay rapidly. Consequently, the error $\sum_{t=1}^{T} \sigma_t^2$ would be negligibly small when $r_2$ exceeds the intrinsic rank of the gradients. By maintaining low-rank sketch matrix $S_{\tau, T}$, we reduce the cost of computing $\sum_{t=1}^{T} G_{\tau,t}^\prime G_{\tau,t}^{\prime\top}$ from $O(m^2nT)$ to $O(mnr_2T)$, where $r_2 \ll m$. Proposition~\ref{thm:low_rank2} ensure that our low-rank approximation captures the most significant directions in the gradient space. The error bound provides practical guidance for choosing the rank $r_2$: larger values lead to better approximation at the expense of additional computation and memory. To better preserve the task information, we set $r_2$ to be slightly larger than $r_1$, where $r_1$ is the projection rank introduced in Section~\ref{sec:32}. The proof is provided in Appendix~\ref{app:proof2}.
\end{remark*}

\textbf{Update Orthogonal Basis Matrix.} 
Once the final task-specific sketch matrix $S_{\tau,T}$ is computed, we use it to update the orthogonal basis matrix $\mathcal{M}_{\tau-1}$ to incorporate the optimization subspace of the task $\tau$. First, we extract the principal directions of the current task by performing SVD on its sketch matrix:
\begin{equation}
\label{eq:update_space-1}
\begin{aligned}
    U_{\tau}\Sigma_{\tau}V_{\tau}^\top = \text{SVD}(S_{\tau, T}).
\end{aligned}
\end{equation}
Then, we determine the number of directions to retain based on the sum of squared singular values. Following the principle of matrix approximation with SVD, we select $k$ as the biggest value that satisfies:
\begin{equation}
\label{eq:update_space-2}
    \frac{\sum_{i=1}^k \sigma_i^2}{\sum_{j=1}^{r_2} \sigma_j^2} \leq \epsilon_{th},
\end{equation}
where $\epsilon_{th} \in (0,1]$ is a threshold hyperparameter controlling the ratio to preserve, and $\sigma_i$ is the $i$-th singular values in descending order. This criterion ensures that the selected $k$ directions capture at least $\epsilon_{th}$ fraction of the total variance in the gradient space. Finally, we expand the orthogonal basis matrix $\mathcal{M}_{\tau-1}$ by incorporating these new directions:
\begin{equation}
\label{eq:update_space-3}
    \mathcal{M}_{\tau} = [\mathcal{M}_{\tau-1}, U_{\tau}[:,:k]].
\end{equation}
The above selection and update process ensure that we capture the most important learning directions for each task while maintaining orthogonality between different tasks' subspaces.

Due to space constraints, the complete CoSO algorithm is presented in Appendix~\ref{app:algo}.

\section{Experiments}
\label{sec:exp}
    We conduct comprehensive experiments with varying numbers of sequential tasks to evaluate CoSO's effectiveness across multiple datasets. We first outline our experimental settings, then present detailed results and analyses.

\subsection{Experimental Settings}

    \textbf{Datasets and Evaluation Metrics.} Following previous works \citep{wang2022learning,Liang2024inflora}, we evaluate CoSO on three widely-used continual learning benchmarks: ImageNet-R \citep{Hendrycks2021imagenet}, CIFAR100 \citep{krizhevsky2009cifar}, and DomainNet \citep{peng2019domainnet}. ImageNet-R contains 200 classes from ImageNet with artistic style variations. Similar to existing works \citep{Smith2023coda,Liang2024inflora,wu2025sdlora}, we create three different splits of ImageNet-R: 5 tasks with 40 classes per task, 10 tasks with 20 classes per task, and 20 tasks with 10 classes per task. For CIFAR100, we divide it into 10 tasks, each containing 10 classes. DomainNet consists of 345 classes across six distinct domains and is split into 5 tasks, with 69 classes per task.

    We evaluate our method using two complementary metrics that are widely adopted in existing continual learning methods \citep{wang2022learning,Liang2024inflora,wu2025sdlora}. The first metric is the final accuracy $ACC_T$, which evaluates the model's overall performance across all tasks after the complete training process. The second metric is the average accuracy $\overline{ACC}_T$, which measures the model's learning stability throughout the training sequence and is calculated as $\overline{ACC}_T = \frac{1}{T}\sum_{i=1}^T ACC_i$, where $T$ denotes the total number of tasks. These two metrics capture both the model's ability to learn new tasks and retain knowledge of previously learned tasks, providing a comprehensive assessment of continual learning performance.

    \textbf{Baselines and Implementation Details.} We compare CoSO with several state-of-the-art PEFT-based methods: L2P \citep{wang2022learning}, DualPrompt \citep{wang2022dualprompt}, CODA-Prompt (CODA-P) \citep{Smith2023coda}, InfLoRA \citep{Liang2024inflora}, $\text{VPT-NSP}^2$ \citep{lu2024visual}, and SD-LoRA \citep{wu2025sdlora}. Comparing against both prompt-based and LoRA-based methods allows us to comprehensively evaluate the effectiveness of CoSO. In addition to the ViT-B/16 \citep{dosovitskiy2021vit} pretrained on ImageNet-1K, we also evaluate a self-supervised ViT-B/16 obtained with DINO \citep{caron2021emerging}. Details of the experimental setup are provided in Appendix~\ref{app:hyperparam}.

    \begin{table}[t]
        \caption{Results (\%) on ImageNet-R with varying numbers of tasks (5, 10 and 20). All reported results with mean and standard deviation are computed over 3 independent runs.}
        \label{tab:imgnet-r}
        \centering
        \begin{tabular}{lllllll}
            \toprule
            \multirow{2}{*}{Method} & \multicolumn{2}{c}{ImageNet-R (5 Tasks)} & \multicolumn{2}{c}{ImageNet-R (10 Tasks)} & \multicolumn{2}{c}{ImageNet-R (20 Tasks)} \\
            \cmidrule(r){2-3} \cmidrule(r){4-5} \cmidrule(r){6-7}
            & $ACC_5$ & $\overline{ACC}_5$ & $ACC_{10}$ & $\overline{ACC}_{10}$ & $ACC_{20}$ & $\overline{ACC}_{20}$ \\
            \midrule
            L2P & $65.03_{\pm 0.03}$  & $69.97_{\pm 0.15}$ & $62.87_{\pm 0.72}$ & $68.90_{\pm 0.58}$ & $58.64_{\pm 0.34}$ & $65.57_{\pm 0.35}$ \\
            DualPrompt & $68.24_{\pm 0.23}$  & $71.82_{\pm 0.39}$ & $65.30_{\pm 0.52}$ & $69.62_{\pm 0.29}$ & $60.47_{\pm 0.54}$ & $65.91_{\pm 0.52}$ \\
            CODA-P & $73.65_{\pm 0.15}$  & $77.88_{\pm 0.30}$ & $72.10_{\pm 0.29}$ & $76.90_{\pm 0.41}$ & $67.16_{\pm 0.11}$ & $72.34_{\pm 0.44}$ \\
            InfLoRA & $77.53_{\pm 0.30}$  & $82.24_{\pm 0.11}$ & $74.43_{\pm 0.31}$ & $80.50_{\pm 0.06}$ & $70.30_{\pm 0.14}$ & $77.04_{\pm 0.06}$ \\
            SD-LoRA & $79.15_{\pm 0.20}$ & $83.01_{\pm 0.42}$ & $77.34_{\pm 0.35}$ & $82.04_{\pm 0.24}$ & $75.26_{\pm 0.37}$ & $80.22_{\pm 0.72}$\\
            $\text{VPT-NSP}^2$ & $79.72_{\pm 0.19}$  & $84.33_{\pm 0.29}$ & $77.87_{\pm 0.10}$ & $83.09_{\pm 0.26}$ & $75.42_{\pm 0.27}$ & $81.32_{\pm 0.21}$ \\
            CoSO & $\textbf{82.10}_{\pm 0.13}$ & $\textbf{86.38}_{\pm 0.07}$ & $\textbf{81.10}_{\pm 0.39}$ & $\textbf{85.56}_{\pm 0.13}$ & $\textbf{78.19}_{\pm 0.28}$ & $\textbf{83.69}_{\pm 0.12}$ \\
            \bottomrule
        \end{tabular}
    \end{table}

    \begin{figure}[t]
        \centering
        \begin{subfigure}{0.325\textwidth}
            \centering
            \includegraphics[width=\linewidth]{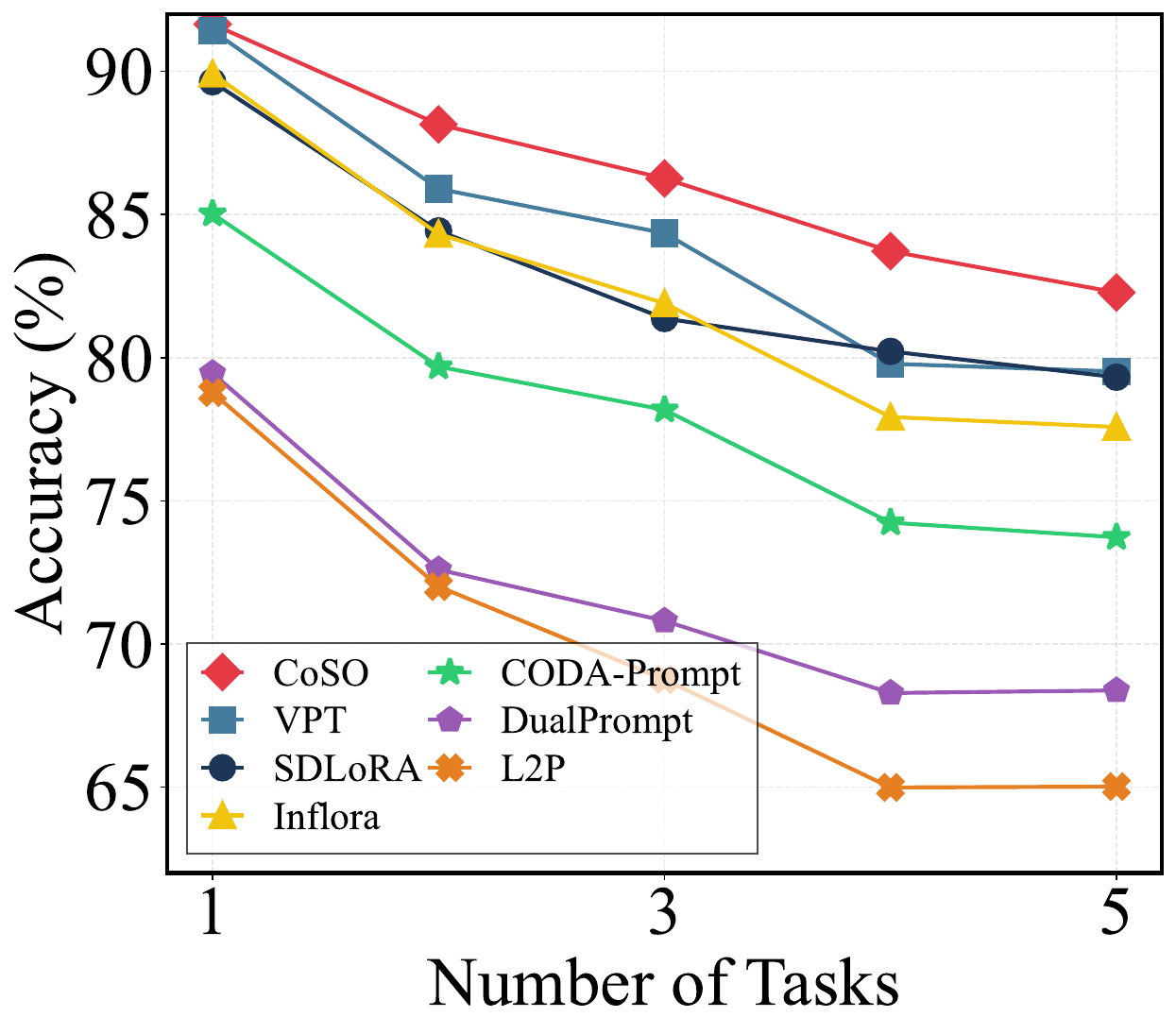}
            \text{(a) 5 tasks}
            \label{fig:sub1}
        \end{subfigure}
        \hfill
        \begin{subfigure}{0.325\textwidth}
            \centering
            \includegraphics[width=\linewidth]{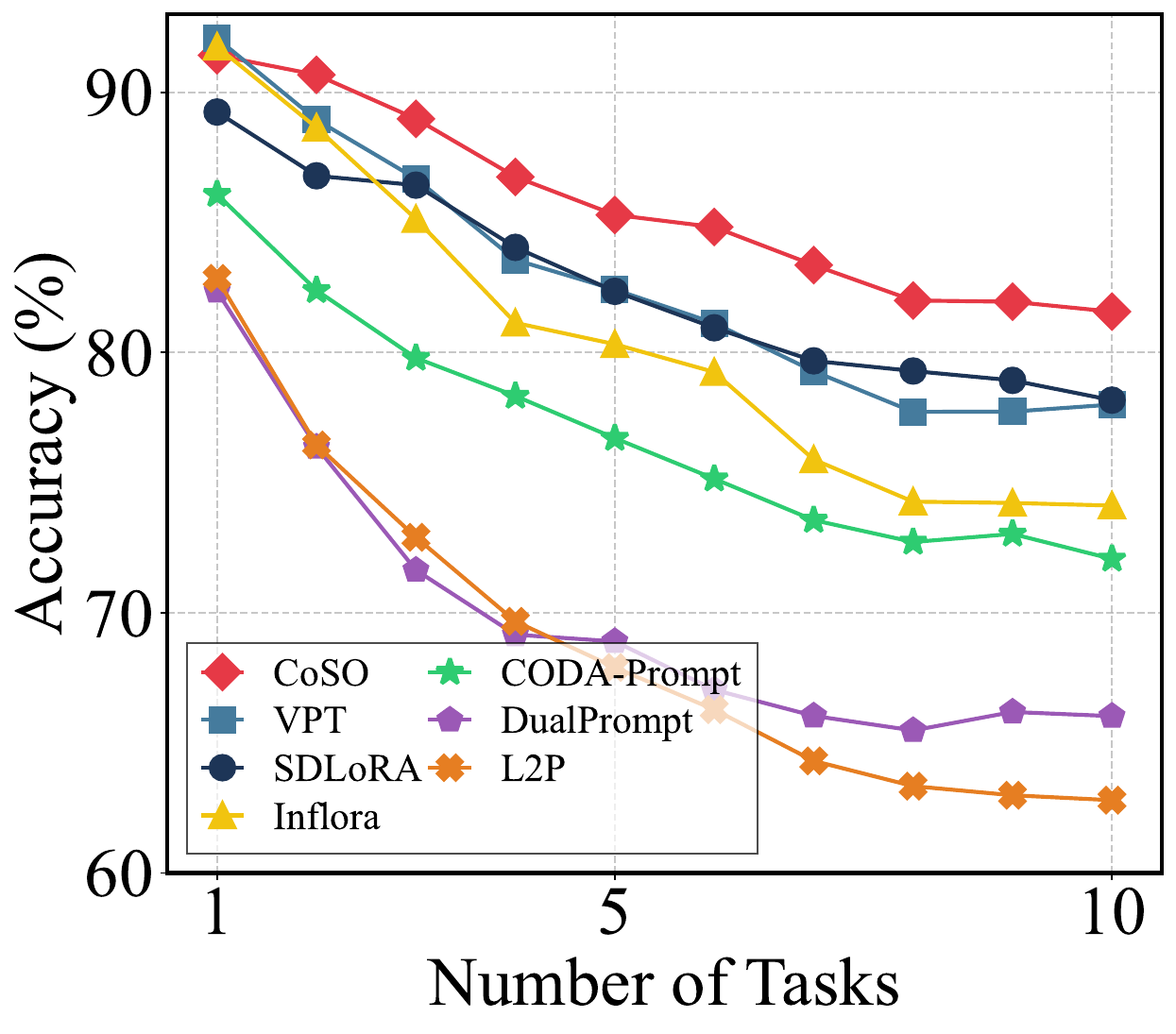}
            \text{(b) 10 tasks}
            \label{fig:sub2}
        \end{subfigure}
        \hfill
        \begin{subfigure}{0.325\textwidth}
            \centering
            \includegraphics[width=\linewidth]{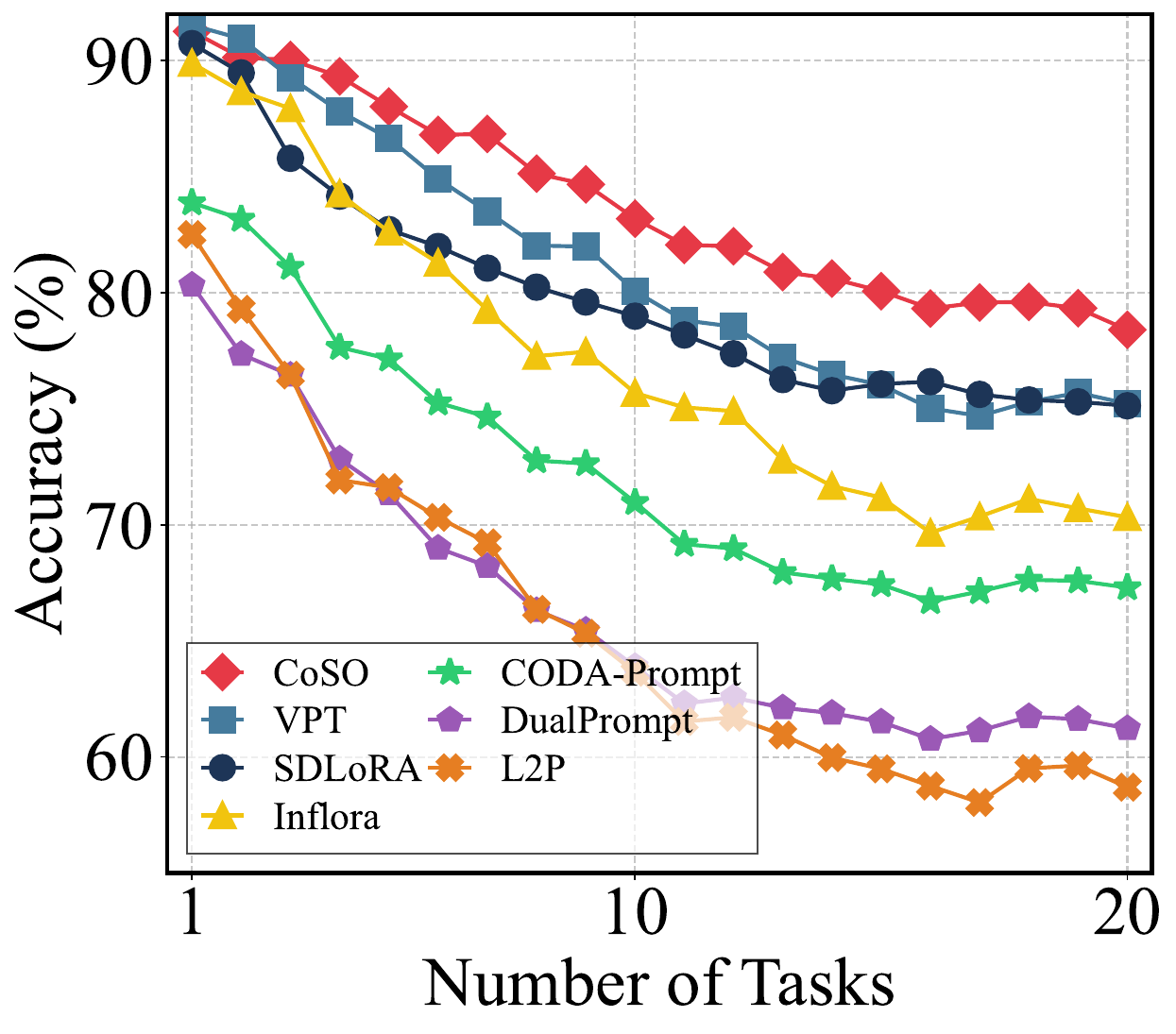}
            \text{(c) 20 tasks}
            \label{fig:sub3}
        \end{subfigure}
        \caption{The detailed performance during the learning of ImageNet-R on (a) 5 tasks, (b) 10 tasks, and (c) 20 tasks.}
        \label{fig:subfig_example}
    \end{figure}

\subsection{Experimental Results}

    We evaluate CoSO against state-of-the-art continual learning methods across different experimental settings. Table~\ref{tab:imgnet-r} shows the performance comparison on ImageNet-R under various task partitions (5, 10, and 20 tasks). Across all partitions, CoSO delivers the highest final accuracy ($ACC$) and average accuracy ($\overline{ACC}$), confirming its robustness to mitigate forgetting. For the most challenging setting (20 tasks), CoSO attains 78.19\% final accuracy and 83.69\% average accuracy, while the best baseline method achieves 75.42\% and 81.32\%, respectively. For the ImageNet-R 10 tasks scenario, CoSO improves the final accuracy by 3.23\% and the average accuracy by 2.47\% compared to the best baseline method. Likewise, in the 5 tasks setting, CoSO still leads by 2.38\% in final accuracy and 2.05\% in average accuracy. This margin highlights CoSO's exceptional resistance to forgetting and its strong capacity to integrate new knowledge without eroding prior learning.
    
    Figure~\ref{fig:subfig_example} illustrates the evolution of accuracy throughout the continual learning process for various methods evaluated on ImageNet-R. It is evident that CoSO consistently maintains superior performance relative to other approaches, both during the intermediate phases and at the end of training. This ongoing superiority underscores CoSO's effectiveness in reducing interference from newly introduced tasks, resulting in a significantly slower decline in accuracy compared to competing methods. Complementary results in Table~\ref{tab:cifar-domain} reveal the same trend on CIFAR100 and DomainNet. On the DomainNet benchmark, CoSO outperforms the best baseline method by 1.75\% in final accuracy and 1.37\% in average accuracy, confirming its ability to generalize across heterogeneous visual domains. A detailed analysis of computational and memory costs are presented in Appendix~\ref{app:memory}. The additional results with DINO \citep{caron2021emerging} are provided in Appendix~\ref{app:dino}.

    \begin{table}[t]
        \caption{Results (\%) on CIFAR100 (10 Tasks) and DomainNet (5 Tasks). All reported results with mean and standard deviation are computed over 3 independent runs.}
        \label{tab:cifar-domain}
        \centering
        \begin{tabular}{lllll}
            \toprule
            \multirow{2}{*}{Method} & \multicolumn{2}{c}{CIFAR100 (10 Tasks)} & \multicolumn{2}{c}{DomainNet (5 Tasks)}  \\
            \cmidrule(r){2-3} \cmidrule(r){4-5} 
            & $ACC_{10}$ & $\overline{ACC}_{10}$ & $ACC_5$ & $\overline{ACC}_5$ \\
            \midrule
            L2P & $82.64_{\pm 0.26}$ & $87.90_{\pm 0.19}$ & $70.03_{\pm 0.09}$ & $75.65_{\pm 0.06}$ \\
            DualPrompt & $84.68_{\pm 0.22}$ & $90.12_{\pm 0.05}$ & $72.25_{\pm 0.05}$ & $77.84_{\pm 0.02}$ \\
            CODA-P & $86.60_{\pm 0.37}$ & $91.46_{\pm 0.20}$ & $73.16_{\pm 0.07}$ & $78.75_{\pm 0.04}$ \\
            InfLoRA & $86.85_{\pm 0.08}$ & $91.45_{\pm 0.16}$ & $73.09_{\pm 0.11}$ & $79.21_{\pm 0.08}$ \\
            SD-LoRA & $87.30_{\pm 0.45}$ & $91.81_{\pm 0.27}$ & $73.20_{\pm 0.12}$ & $79.03_{\pm 0.04}$ \\
            $\text{VPT-NSP}^2$ & $88.09_{\pm 0.12}$ & $92.48_{\pm 0.11}$ & $72.52_{\pm 0.13}$ & $78.68_{\pm 0.06}$ \\
            CoSO & $\textbf{88.77}_{\pm 0.16}$ & $\textbf{92.99}_{\pm 0.23}$ & $\textbf{74.27}_{\pm 0.07}$ & $\textbf{80.05}_{\pm 0.04}$ \\
            \bottomrule
        \end{tabular}
    \end{table}

    \textbf{Ablation Study.} We conduct comprehensive ablation studies on ImageNet-R benchmark to validate the individual contributions of the orthogonal projection mechanism and the Frequent Directions (FD) based subspace consolidation. Specifically, we compare CoSO with two variants. The first variant (w/o Orth) removes the orthogonal projection, which directly uses the original gradients $G_{\tau,t}$ for optimization instead of the orthogonally projected gradients $G_{\tau,t}^\prime$. This variant optimizes parameters in continuous subspaces without any orthogonality constraint, thereby ignoring task interference. The second variant (w/o FD) retains orthogonality but, instead of employing FD to consolidate all intermediate gradients from the current task, constructs the task-specific subspace using only the final subspace obtained at the end of that task.

    The results are summarized in Table~\ref{tab:ablation}. Eliminating orthogonal projection (w/o Orth) leads to a sharp performance drop (8.52\% in final accuracy) on 20 Tasks setting, highlighting the importance of excluding new gradients from the historical subspace to prevent interference. Replacing FD with the simplified strategy that builds each task-specific subspace from only the final gradient subspace (w/o FD) also degrades performance, lowering final accuracy by 1.65\%, 1.89\% and 1.59\% for 5, 10 and 20 Tasks settings, respectively. This drop confirms that aggregating all intermediate gradients through incremental FD updates captures richer task information than using a single terminal subspace. Across the table, the full method delivers the highest final and average accuracies, indicating that both orthogonal projection and FD consolidation are indispensable for robust continual learning.

\begin{table}[t]
\caption{Ablation study results (\%) on ImageNet-R with varying numbers of tasks (5, 10 and 20).}
\label{tab:ablation}
\centering
\begin{tabular}{lllllll}
    \toprule
    \multirow{2}{*}{Method} & \multicolumn{2}{c}{ImageNet-R (5 Tasks)} & \multicolumn{2}{c}{ImageNet-R (10 Tasks)} & \multicolumn{2}{c}{ImageNet-R (20 Tasks)} \\
    \cmidrule(r){2-3} \cmidrule(r){4-5} \cmidrule(r){6-7}
    & $ACC_5$ & $\overline{ACC}_5$ & $ACC_{10}$ & $\overline{ACC}_{10}$ & $ACC_{20}$ & $\overline{ACC}_{20}$ \\
    \midrule
    w/o Orth & $79.35$ & $85.22$ & $75.90$ & $83.43$ & $69.75$ & $78.88$ \\
    w/o FD   & $80.72$ & $85.44$ & $78.83$ & $84.45$ & $76.68$ & $82.41$ \\
    CoSO     & $82.37$ & $86.46$ & $80.72$ & $85.67$ & $78.27$ & $83.62$ \\
    \bottomrule
\end{tabular}
\end{table}

\section{Conclusion}
\label{sec:conclusion}
In this paper, we propose Continuous Subspace Optimization for Continual Learning (CoSO). CoSO optimizes the pre-trained models within continuous subspaces. By maintaining orthogonality between the current task's optimization subspace and that of historical tasks, CoSO effectively mitigates the interference. CoSO maintains a compact task-specific component while learning a task. After completing the current task, the task-specific component is used to update the historical task subspace. Extensive experiments on standard benchmarks demonstrate that CoSO consistently outperforms state-of-the-art baselines in both final accuracy and average accuracy over time, confirming its effectiveness and robustness across diverse data streams. In the future, a challenging open problem is to extend CoSO to multimodal task settings.

%%%%%%%%%%%%%%%%%%%%%%%%%%%%%%%%%%%%%%%%%%%%%%%%%%%%%%%%%%%%
\section*{Acknowledgments and Disclosure of Funding}

This work was partially supported by National Science and Technology Major Project (2022ZD0114801), and NSFC (U23A20382).

%%%%%%%%%%%%%%%%%%%%%%%%%%%%%%%%%%%%%%%%%%%%%%%%%%%%%%%%%%%%
% \section*{References}

\bibliography{neurips_2025}
\bibliographystyle{plainnat}

%%%%%%%%%%%%%%%%%%%%%%%%%%%%%%%%%%%%%%%%%%%%%%%%%%%%%%%%%%%%

\newpage
\section*{NeurIPS Paper Checklist}

\begin{enumerate}

\item {\bf Claims}
    \item[] Question: Do the main claims made in the abstract and introduction accurately reflect the paper's contributions and scope?
    \item[] Answer: \answerYes{} % Replace by \answerYes{}, \answerNo{}, or \answerNA{}.
    \item[] Justification: Please refer to Section~\ref{sec:method} and \ref{sec:exp}.

\item {\bf Limitations}
    \item[] Question: Does the paper discuss the limitations of the work performed by the authors?
    \item[] Answer: \answerYes{} % Replace by \answerYes{}, \answerNo{}, or \answerNA{}.
    \item[] Justification: Please refer to Section~\ref{sec:conclusion}.

\item {\bf Theory assumptions and proofs}
    \item[] Question: For each theoretical result, does the paper provide the full set of assumptions and a complete (and correct) proof?
    \item[] Answer: \answerYes{} % Replace by \answerYes{}, \answerNo{}, or \answerNA{}.
    \item[] Justification: Please refer to Section~\ref{sec:method} for the assumptions. Please refer to Appendixes for the complete proof.

    \item {\bf Experimental result reproducibility}
    \item[] Question: Does the paper fully disclose all the information needed to reproduce the main experimental results of the paper to the extent that it affects the main claims and/or conclusions of the paper (regardless of whether the code and data are provided or not)?
    \item[] Answer: \answerYes{} % Replace by \answerYes{}, \answerNo{}, or \answerNA{}.
    \item[] Justification: Please refer to Section~\ref{sec:exp} and Appendix~\ref{app:hyperparam}.

\item {\bf Open access to data and code}
    \item[] Question: Does the paper provide open access to the data and code, with sufficient instructions to faithfully reproduce the main experimental results, as described in supplemental material?
    \item[] Answer: \answerYes{} % Replace by \answerYes{}, \answerNo{}, or \answerNA{}.
    \item[] Justification: Please refer to the supplemental materials.
    \item[] Guidelines:
    \begin{itemize}
        \item The answer NA means that paper does not include experiments requiring code.
        \item Please see the NeurIPS code and data submission guidelines (\url{https://nips.cc/public/guides/CodeSubmissionPolicy}) for more details.
        \item While we encourage the release of code and data, we understand that this might not be possible, so "No" is an acceptable answer. Papers cannot be rejected simply for not including code, unless this is central to the contribution (e.g., for a new open-source benchmark).
        \item The instructions should contain the exact command and environment needed to run to reproduce the results. See the NeurIPS code and data submission guidelines (\url{https://nips.cc/public/guides/CodeSubmissionPolicy}) for more details.
        \item The authors should provide instructions on data access and preparation, including how to access the raw data, preprocessed data, intermediate data, and generated data, etc.
        \item The authors should provide scripts to reproduce all experimental results for the new proposed method and baselines. If only a subset of experiments are reproducible, they should state which ones are omitted from the script and why.
        \item At submission time, to preserve anonymity, the authors should release anonymized versions (if applicable).
        \item Providing as much information as possible in supplemental material (appended to the paper) is recommended, but including URLs to data and code is permitted.
    \end{itemize}

\item {\bf Experimental setting/details}
    \item[] Question: Does the paper specify all the training and test details (e.g., data splits, hyperparameters, how they were chosen, type of optimizer, etc.) necessary to understand the results?
    \item[] Answer: \answerYes{} % Replace by \answerYes{}, \answerNo{}, or \answerNA{}.
    \item[] Justification: Please refer to Section~\ref{sec:exp} and Appendix~\ref{app:hyperparam}.

\item {\bf Experiment statistical significance}
    \item[] Question: Does the paper report error bars suitably and correctly defined or other appropriate information about the statistical significance of the experiments?
    \item[] Answer: \answerYes{} % Replace by \answerYes{}, \answerNo{}, or \answerNA{}.
    \item[] Justification: We report the main results with mean and standard deviation, which are computed over 3 independent runs.

\item {\bf Experiments compute resources}
    \item[] Question: For each experiment, does the paper provide sufficient information on the computer resources (type of compute workers, memory, time of execution) needed to reproduce the experiments?
    \item[] Answer: \answerYes{} % Replace by \answerYes{}, \answerNo{}, or \answerNA{}.
    \item[] Justification: Please refer to Section~\ref{sec:exp} and Appendix~\ref{app:hyperparam}.
    
\item {\bf Code of ethics}
    \item[] Question: Does the research conducted in the paper conform, in every respect, with the NeurIPS Code of Ethics \url{https://neurips.cc/public/EthicsGuidelines}?
    \item[] Answer: \answerYes{} % Replace by \answerYes{}, \answerNo{}, or \answerNA{}.
    \item[] Justification: The authors have read the NeurIPS Code of Ethics and ensured that our research conforms to it.

\item {\bf Broader impacts}
    \item[] Question: Does the paper discuss both potential positive societal impacts and negative societal impacts of the work performed?
    \item[] Answer: \answerNA{} % Replace by \answerYes{}, \answerNo{}, or \answerNA{}.
    \item[] Justification: This paper is about continual learning and does not involve societal impact.
    
\item {\bf Safeguards}
    \item[] Question: Does the paper describe safeguards that have been put in place for responsible release of data or models that have a high risk for misuse (e.g., pretrained language models, image generators, or scraped datasets)?
    \item[] Answer: \answerNA{} % Replace by \answerYes{}, \answerNo{}, or \answerNA{}.
    \item[] Justification: This paper is about continual learning and does not have a risk for misuse.

\item {\bf Licenses for existing assets}
    \item[] Question: Are the creators or original owners of assets (e.g., code, data, models), used in the paper, properly credited and are the license and terms of use explicitly mentioned and properly respected?
    \item[] Answer: \answerYes{} % Replace by \answerYes{}, \answerNo{}, or \answerNA{}.
    \item[] Justification: We have properly cited all data, code, and models used in this paper.

\item {\bf New assets}
    \item[] Question: Are new assets introduced in the paper well documented and is the documentation provided alongside the assets?
    \item[] Answer: \answerNA{} % Replace by \answerYes{}, \answerNo{}, or \answerNA{}.
    \item[] Justification: The paper does not release new assets.

\item {\bf Crowdsourcing and research with human subjects}
    \item[] Question: For crowdsourcing experiments and research with human subjects, does the paper include the full text of instructions given to participants and screenshots, if applicable, as well as details about compensation (if any)? 
    \item[] Answer: \answerNA{} % Replace by \answerYes{}, \answerNo{}, or \answerNA{}.
    \item[] Justification: The paper does not involve crowdsourcing nor research with human subjects.

\item {\bf Institutional review board (IRB) approvals or equivalent for research with human subjects}
    \item[] Question: Does the paper describe potential risks incurred by study participants, whether such risks were disclosed to the subjects, and whether Institutional Review Board (IRB) approvals (or an equivalent approval/review based on the requirements of your country or institution) were obtained?
    \item[] Answer: \answerNA{} % Replace by \answerYes{}, \answerNo{}, or \answerNA{}.
    \item[] Justification: The paper does not involve crowdsourcing nor research with human subjects.

\item {\bf Declaration of LLM usage}
    \item[] Question: Does the paper describe the usage of LLMs if it is an important, original, or non-standard component of the core methods in this research? Note that if the LLM is used only for writing, editing, or formatting purposes and does not impact the core methodology, scientific rigorousness, or originality of the research, declaration is not required.
    %this research? 
    \item[] Answer: \answerNA{} % Replace by \answerYes{}, \answerNo{}, or \answerNA{}.
    \item[] Justification: This research does not involve LLMs as any important, original, or non-standard components.

\end{enumerate}

%%%%%%%%%%%%%%%%%%%%%%%%%%%%%%%%%%%%%%%%%%%%%%%%%%%%%%%%%%%%
\newpage
\appendix

\section{Proof of Proposition~\ref{thm:low_rank2}}
\label{app:proof2}

    Recall that $\{G'_{\tau,t}\}_{t=1}^{T}\subset\mathbb{R}^{m\times n}$ is the sequence of projected
    gradients for task $\tau$, $A_t = G_{\tau, t}^\prime G_{\tau, t}^{\prime\top}$, $A=\sum_{t=1}^T G_{\tau,t}^\prime G_{\tau,t}^{\prime\top}$, $\tilde{A_t} = Q_{\tau,t}Q_{\tau,t}^\top$, $\tilde{A} = \sum_{t=1}^T Q_{\tau,t}Q_{\tau,t}^\top$ and sketch matrix $S_{\tau, T}\in\mathbb{R}^{m\times r_2}$.
    
    Because $Q_{\tau,t}$ is the rank $r_2$ approximation of $G_{\tau,t}^\prime$, for every step $t$, we have
    \begin{equation}
    \label{eq:single_step}
      \|A_t-\tilde{A_t}\|_{2} = \sigma_t^2,
    \end{equation}
    where $\sigma_t$ is the $(r_2+1)$ singular value of $G_{\tau,t}^\prime$.

    Using the triangle inequality together with Eq.~(\ref{eq:single_step}),
    \begin{equation}
    \label{eq:sum_before_fd}
      \|A-\tilde{A}\|_{2} = \left\| \sum_{t=1}^{T}(A_t-\tilde{A_t})\right\|_{2}
      \le \sum_{t=1}^{T} \|A_t-\tilde{A_t}\|_{2}
      = \sum_{t=1}^{T} \sigma_t^2.
    \end{equation}
    
    Since we use FD to compute $S_{\tau, T}$ based on $\{Q_{\tau,t}\}_{t=1}^T$, from Theorem~1.1 of \citet{Ghashami2016fd}, we have
    \begin{equation}
    \label{eq:fd_internal}
      \|\tilde{A}-S_{\tau, T}S_{\tau, T}^{\top}\|_{2}
      \le \frac{\|\tilde A-[\tilde A]_k\|_{F}^{2}}{r_2-k},
    \end{equation}
    where $[\tilde A]_k$ is the minimizer of $\|\tilde A-[\tilde A]_k\|_{F}$ overall rank $k$ matrices.
    Applying the triangle inequality to $\|A-S_{\tau, T}S_{\tau, T}^{\top}\|$ and substituting Eq.~\eqref{eq:sum_before_fd} and \eqref{eq:fd_internal} gives
    \begin{equation}
        \begin{aligned}
            \|A - S_{\tau, T}S_{\tau, T}^{\top}\|_{2} &\le \|A-\tilde{A}\|_{2} + \|\tilde{A}-S_{\tau, T}S_{\tau, T}^{\top}\|_{2} \\
            & \le  \sum_{t=1}^{T} \sigma_t^2 + \frac{\|\tilde A-[\tilde A]_k\|_{F}^{2}}{r_2 - k},
        \end{aligned}
    \end{equation}
    
    which is exactly \eqref{eq:fd_global_bound}.

\section{CoSO Algorithm}
\label{app:algo}

We present the the detailed procedure in Algorithm~\ref{alg:CoSO}.

\begin{algorithm}[t]
    \caption{CoSO for Continual Learning}
    \label{alg:CoSO}
    \begin{algorithmic}[1]
    \State \textbf{Input:} A layer weight matrix $W \in \mathbb{R}^{m \times n}$, step size $\eta$, decay rates $\beta_1, \beta_2$, projection rank $r_1$, FD rank $r_2$, threshold $\epsilon$ and update gap $K$.
    \State Initialize first-order moment $M_0 \in \mathbb{R}^{m \times r} \gets 0$
    \State Initialize second-order moment $V_0 \in \mathbb{R}^{m \times r} \gets 0$
    \State Initialize sketch matrix $S_{\tau,0} \in \mathbb{R}^{m \times r} \gets 0$
    \State Initialize orthogonal projection matrix $\mathcal{M}_0$ $\gets$ 0
    \For {Task $\tau \in 1\dots N$}
        \For {step $t \in 1\dots T$}
            \State $G_{\tau, t} \gets \nabla_{W_{\tau, t}} L(W_{\tau, t})$ \Comment{Compute mini-batch gradient for task $\tau$}
            \State $G_{\tau,t}^\prime \gets G_{\tau, t} - \mathcal{M}_{\tau-1}\mathcal{M}^\top_{\tau-1} G_{\tau, t}$ \Comment{Orthogonal projection}
            \If {$t \bmod K == 0$}
                \State $U\Sigma V^\top = \text{SVD}(G_{\tau, t}^\prime)$
                \State $P_{\tau, t} = U[:,:r_1]$ \Comment{Compute projection matrix $P_{\tau, t}$}
                \State Update $S_{\tau, t}$ through Eq.~(\ref{eq:Q}) and (\ref{eq:S}) \Comment{Use FD to consolidate gradient information}
            \Else
                \State $P_{\tau,t} \gets P_{\tau, t-1}$    
                \State $S_{\tau,t} \gets S_{\tau, t-1}$
            \EndIf
            \State $R_{\tau,t} \gets P_{\tau, t}^\top G_{\tau,t}^\prime$ \Comment{Project orthogonal gradient into low rank space}
            \State Use $R_{\tau,t}$ to compute $N_{\tau,t}$ through Eq.~(\ref{eq:adam-update}) \Comment{Update $R_{\tau,t}$ by Adam}
            \State $\tilde{G}_{\tau,t} \gets P_{\tau,t}N_{\tau,t}$ \Comment{Project gradient back to original space}
            \State $W_{\tau,t} \gets W_{\tau,t-1} - \eta \cdot \tilde{G}_{\tau,t}$
        \EndFor
        \State Update the historical subspaces basis matrix $\mathcal{M}_{\tau-1}$ through Eq.~(\ref{eq:update_space-1}), (\ref{eq:update_space-2}) and (\ref{eq:update_space-3})
    \EndFor
    \end{algorithmic}
\end{algorithm}

\section{Experimental Setups and Implementation Details}
\label{app:hyperparam}

Following existing works \citep{Smith2023coda,wu2025sdlora}, we adopt ViT-B/16 \citep{dosovitskiy2021vit} pre-trained on ImageNet-21K and fine-tuned on ImageNet-1K as our backbone model, which consists of 12 transformer blocks. For fair comparison, all methods use the same ViT-B/16 backbone and optimizer. Additionally, we also evaluate a self-supervised ViT-B/16 obtained with DINO \citep{caron2021emerging}. The optimization is performed using Adam \citep{kingma2014adam} optimizer with $\beta_1=0.9$ and $\beta_2=0.999$. The training epochs vary across datasets: 40 epochs for ImageNet-R, 20 epochs for CIFAR100, and 5 epochs for DomainNet. We maintain a consistent batch size of 128 across all experiments. Results are averaged over 3 independent runs, and we report the corresponding standard deviation. Notably, CoSO only optimize the output projection layers in multi-head attention module rather than QKV transformations.

We present the detailed hyperparameter settings of CoSO in Table~\ref{tab:hyperparams}. These hyperparameters are carefully tuned to balance memory efficiency and performance, reflecting the varying complexity of the datasets. The hyperparameter settings of baseline methods are following existing work \citep{wang2022dualprompt,Smith2023coda,Liang2024inflora,lu2024visual,wu2025sdlora}. For all datasets, we employ minimal data augmentation, consisting of random resized cropping to $224\times 224$ pixels and random horizontal flipping during training, without any additional augmentation techniques. To prevent overfitting, we followed $\text{VPT-NSP}^2$ \citep{lu2024visual}, setting the temperature parameter in the cross-entropy loss to 3 for all datasets. All experiments were conducted on NVIDIA A6000 GPUs with 48GB memory using PyTorch 2.5.1.

\begin{table}[t]
\caption{Hyperparameter settings for different datasets.}
\label{tab:hyperparams}
\centering
\begin{tabular}{llll}
\toprule
Hyperparameter & CIFAR100 & ImageNet-R & DomainNet \\
\midrule
Projection rank ($r_1$) & 15 & 50 & 70 \\
Frequent directions rank ($r_2$) & 100 & 120 & 160 \\
Update gap ($K$) & 1 & 1 & 20 \\
Threshold ($\epsilon_{th}$) & 0.98 & 0.98 & 0.98 \\
\bottomrule
\end{tabular}
\end{table}

The projection rank ($r_1$) determines the dimensionality of the low-rank subspace for gradient projection. For simpler datasets like CIFAR100, a lower value of $r_1 = 15$ is sufficient, while more complex datasets such as ImageNet-R and DomainNet require higher values ($r_1$ = 50 and $r_1$ = 70, respectively) to capture a richer set of gradient directions. The Frequent Directions rank ($r_2$) is consistently set higher than $r_1$ across all datasets. This design choice ensures that CoSO can capture a broader range of directions, reducing information loss during continual learning. As the dataset complexity increases, $r_2$ is adjusted upward to retain more task information.

The update gap K is adjusted based on the characteristics of each dataset. For DomainNet, we use a larger update gap ($K = 20$) due to its larger and more diverse task structure, where frequent updates may become redundant. In contrast, CIFAR100 and ImageNet-R exhibit rapid gradient changes, necessitating a smaller $K$. Finally, the threshold ($\epsilon$) is uniformly set to 0.98 across all datasets. This value is selected to maintain a high retention rate of gradient information within the subspace.

\section{Analysis of Computational and Memory Costs}
\label{app:memory}

\begin{table}[t]
    \caption{Comparison on ImageNet-R (10 Tasks) in terms of computation (GFLOPs) and
    memory usage.}
    \label{tab:memory}
    \centering
    \begin{tabular}{lll}
        \toprule
        Method & GFLOPs & Memory Usage (G)  \\
        \midrule
        L2P & 70.24 & 12.90\\
        DualPrompt & 70.24 & 12.96\\
        CODA-P & 70.24 & 12.97 \\
        InfLoRA & 35.12 & 13.44\\
        SD-LoRA & 35.12 & 15.62 \\
        $\text{VPT-NSP}^2$ & 35.83 & 11.54 \\
        CoSO & 35.12 &  13.61\\
        \bottomrule
    \end{tabular}
\end{table}

We conducted a comparative analysis of CoSO and baseline methods with respect to computational cost (reported as estimated GFLOPs) and memory usage, as summarized in Table \ref{tab:memory}. CoSO requires half the computational cost of prompt-based methods (such as L2P, DualPrompt, and CODA-P), as it avoids the need for twice forward passes through the network. In terms of memory usage, CoSO is on par with other low-rank adaptation techniques such as InfLoRA (13.44). Its slightly higher memory footprint (13.61) stems from using a larger rank for gradient subspace approximation, which enables better capture of task-specific patterns and leads to superior performance. Notably, simply increasing the rank for InfLoRA would not yield similar improvements, as its performance is limited by the constraint of fixed subspaces. Compared with SD-LoRA (15.62), which incurs the greatest memory overhead, CoSO offers a more efficient alternative while delivering competitive performance. Overall, these results highlight CoSO's ability to strike a favorable balance between computational efficiency and memory usage, making it a scalable solution for continual learning across diverse tasks.

\section{Additional Experiment Results on ImageNet-R}
\label{app:dino}

To further verify CoSO's generality, we test it on a self-supervised ViT-B/16 backbone trained with DINO \citep{caron2021emerging} on ImageNet-R (10 Tasks). The results are presented in Table~\ref{tab:dino}. CoSO outperforms the best baseline method with a considerable margin, confirming its ability to generalize across various vision transformers.

\begin{table}[t]
    \caption{Results (\%) on ImageNet-R (10 Tasks). All reported results with mean and standard deviation are computed over 3 independent runs.}
    \label{tab:dino}
    \centering
    \begin{tabular}{lll}
        \toprule
         \multirow{2}{*}{Method (DINO)} & \multicolumn{2}{c}{ImageNet-R (10 Tasks)}  \\
        \cmidrule(r){2-3}  
        & $ACC_{10}$ & $\overline{ACC}_{10}$  \\
        \midrule
        L2P & $61.94_{\pm 0.45}$ & $68.77_{\pm 0.27}$  \\
        DualPrompt & $60.40_{\pm 0.18}$ & $67.65_{\pm 0.07}$  \\
        CODA-P & $64.63_{\pm 0.33}$ & $72.20_{\pm 0.30}$  \\
        InfLoRA & $67.91_{\pm 0.23}$ & $76.40_{\pm 0.03}$  \\
        SD-LoRA & $69.78_{\pm 0.63}$ & $65.73_{\pm 0.35}$  \\
        $\text{VPT-NSP}^2$ & $69.68_{\pm 0.20}$ & $77.24_{\pm 0.16}$  \\
        CoSO & $\textbf{71.60}_{\pm 0.44}$ & $\textbf{79.28}_{\pm 0.16}$ \\
        \bottomrule
    \end{tabular}
\end{table}

\end{document}